\documentclass[sigconf]{acmart}

\AtBeginDocument{%
  }

\copyrightyear{2026}
\acmYear{2026}
\setcopyright{cc}
\setcctype{by}
\acmConference[CHI EA '26]{Extended Abstracts of the 2026 CHI Conference on Human Factors in Computing Systems}{April 13--17, 2026}{Barcelona, Spain}
\acmBooktitle{Extended Abstracts of the 2026 CHI Conference on Human Factors in Computing Systems (CHI EA '26), April 13--17, 2026, Barcelona, Spain}
\acmDOI{10.1145/3772363.3798741}
\acmISBN{979-8-4007-2281-3/2026/04}

\usepackage{booktabs} % For professional-looking tables
\usepackage{caption}  % For table captions
\usepackage{tabularx} % For auto-wrapping text columns
\usepackage[T1]{fontenc}
\usepackage{enumitem}
\usepackage{graphicx}
\usepackage{epstopdf}
\usepackage{multirow}
\usepackage{subcaption}
\usepackage{xcolor}
\usepackage{siunitx}
\usepackage{hyperref}
\usepackage{makecell} 
\usepackage{cuted}
\usepackage{stfloats}

\newcommand{\yr}[1]{#1}
\newcommand{\inj}[1]{#1}

\begin{document}

% \title{PedaCo-Gen: A Pedagogical Framework for Human-AI Collaborative Video Authoring}
% \title{PedaCo-Gen: A Pedagogical Human-AI Collaborative Video Generation System}
\title{PedaCo-Gen: Scaffolding Pedagogical Agency in Human-AI Collaborative Video Authoring}

\author{Injun Baek}
\authornotemark[1]
\affiliation{%
  \institution{Seoul National University}
  \city{Seoul}
  \country{Republic of Korea}
}
\affiliation{%
  \institution{Samsung Electronics}
  \city{Suwon} % 도시는 대표 도시 하나만 적거나 생략 가능
  \country{Republic of Korea}
}
\email{jjune1416@snu.ac.kr}

\author{Yearim Kim}
\authornote{Both authors contributed equally to this research.}
\affiliation{%
  \institution{Seoul National University}
  \city{Seoul}
  \country{Republic of Korea}
}
\email{yerim1656@snu.ac.kr}

\author{Nojun Kwak}
\authornote{Corresponding author}
\affiliation{%
  \institution{Seoul National University}
  \city{Seoul}
  \country{Republic of Korea}
}
\email{nojunk@snu.ac.kr}

\renewcommand{\shortauthors}{Baek et al.}

\begin{abstract} %(T2V)
  While advancements in Text-to-Video generative AI offer a promising path toward democratizing content creation, current models are often optimized for visual fidelity rather than instructional efficacy. This study introduces \textbf{PedaCo-Gen}, a pedagogically-informed human-AI collaborative video generating system for authoring instructional videos based on Mayer’s Cognitive Theory of Multimedia Learning (CTML). Moving away from traditional "one-shot" generation, PedaCo-Gen introduces an Intermediate Representation (IR) phase, enabling educators to interactively review and refine video blueprints—comprising scripts and visual descriptions—with an AI reviewer.
  Our study with 23 education experts demonstrates that PedaCo-Gen significantly enhances video quality across various topics and CTML principles compared to baselines. Participants perceived the AI-driven guidance not merely as a set of instructions but as a metacognitive scaffold that augmented their instructional design expertise, reporting high production efficiency (M=4.26) and guide validity (M=4.04). These findings highlight the importance of reclaiming pedagogical agency through principled co-creation, providing a foundation for future AI authoring tools that harmonize generative power with human professional expertise.
\end{abstract}







\maketitle

\section{Introduction}
\label{sec: intro}
Educational video content has become a fundamental pillar of modern learning, yet the manual production of high-quality instructional materials remains a resource-intensive bottleneck for educators. 
While recent advancements in Text-to-Video (T2V) generative AI~\cite{openai2024sora, google2024veo} offer a promising path toward democratizing content creation, a critical gap persists: existing models are optimized for visual fidelity rather than instructional efficacy.
Current T2V pipelines typically operate as one-shot black boxes~\cite{blackbox2018}, where a prompt is directly converted into a final video.
While prior work has utilized AI for structural augmentations—such as building tutoring interfaces \cite{Calo2024rel2} or adding interactive layers to videos \cite{ALSHAIKH2024rel1}, these supplemental approaches often leave foundational flaws in the core instructional content unaddressed.
Without granular control, educators cannot guarantee pedagogical alignment, leading to videos that inadvertently violate established instructional principles and increase extraneous cognitive load.

To address these challenges, we introduce \textbf{PedaCo-Gen}, a pedagogically informed human-AI collaborative system designed to produce instructionally effective videos. At the core of PedaCo-Gen is the systematic operationalization of \textbf{Mayer’s Cognitive Theory of Multimedia Learning (CTML)}~\cite{mayer2009}. \inj{CTML consists of 12 principles for effective multimedia learning, including the Coherence Principle that reduces extraneous information and the Redundancy Principle that eliminates redundant information (see Appendix~\ref{sec:appendix-ctml} for details). Unlike standard generative pipelines, our system integrates these principles as structured constraints within the generative process.} By introducing an \textbf{Intermediate Representation (IR)} phase—a "video blueprint" consisting of scripts and visual descriptions—PedaCo-Gen allows educators to review and refine content before the final, resource-heavy visual synthesis occurs.

A defining feature of PedaCo-Gen is its \textbf{expert LLM-based review module}, which facilitates a robust human-AI co-creation loop. Rather than serving as a simple automation tool, the system acts as a \textbf{metacognitive scaffold}, providing explainable feedback based on CTML principles. This encourages educators to critically evaluate the AI's suggestions and their own instructional designs. Our findings demonstrate that this "principled friction" does not hinder productivity; in fact, participants rated the system’s production efficiency highly ($M = 4.26$), as it reduced the trial-and-error costs associated with unconstrained generative models.

\begin{figure*}[t]
    \centering
    \includegraphics[width=.95\textwidth]{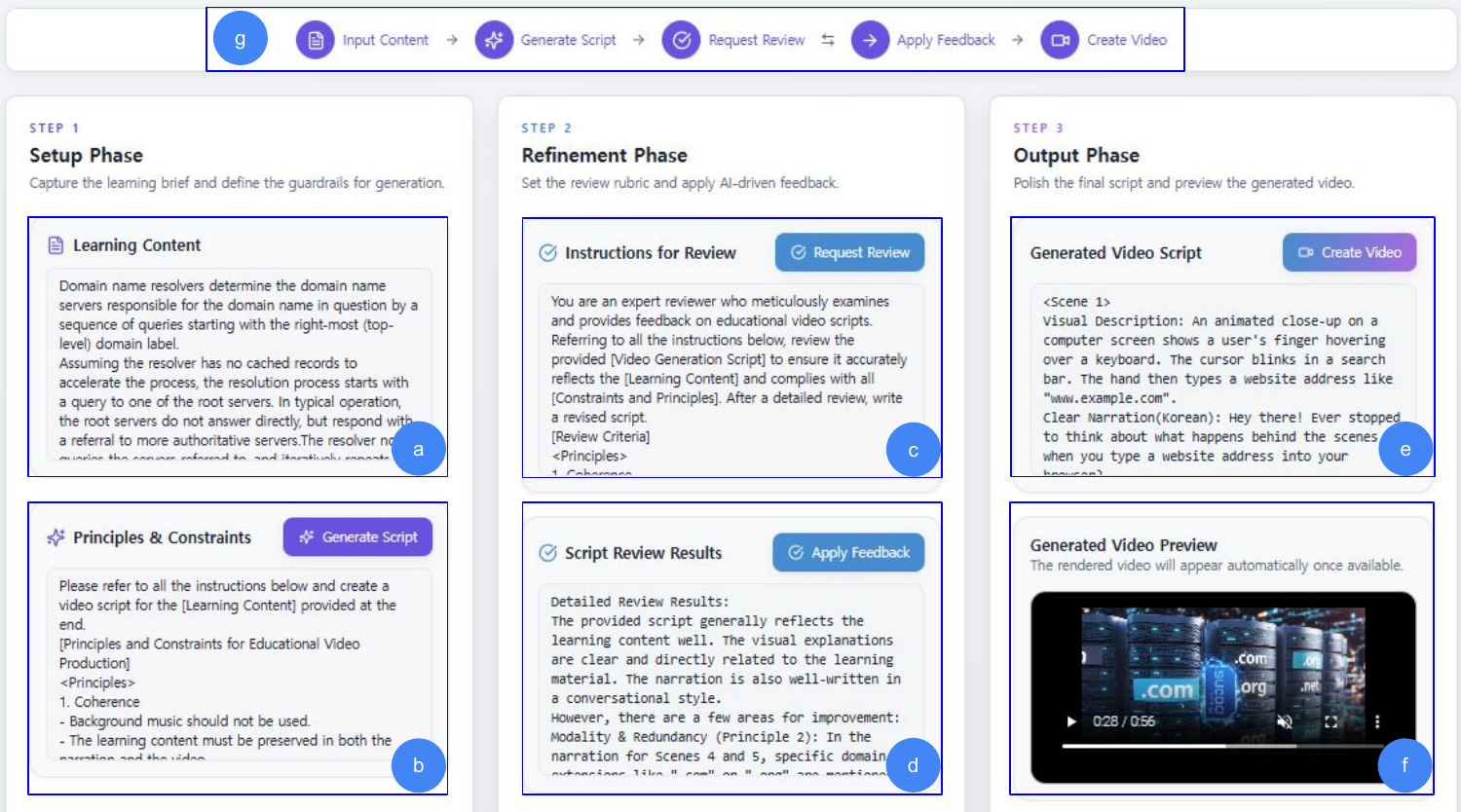}
    \vspace{-1em} 
    \caption{Overview of the PedaCo-Gen system interface for human-AI collaborative educational video authoring. (a) Learning Content Input Panel (b) Principles and Constraints Settings Panel (c) Review Instructions Panel (d) Script Review Results Panel (e) Generated Video Script Panel (f) Generated Video Preview Player (g) Workflow Progress Bar.}
    \vspace{-1em} 
    \label{fig:system}
    \Description{Screenshot of PedaCo-Gen web interface with seven labeled panels. The interface is divided into left and right sections. Left section contains three input panels stacked vertically: (a) a text area for learning content input, (b) a settings panel with checkboxes for CTML principles and constraints, and (c) a text area for custom review instructions. Right section displays four output panels: (d) script review results showing feedback items, (e) the generated video script organized by scenes with visual descriptions and narrations, (f) a video preview player, and (g) a horizontal workflow progress bar at the bottom indicating the current phase.}
\end{figure*}

In this paper, we evaluate PedaCo-Gen through a study with 23 educators. 
It shows that the system significantly enhances the pedagogical quality of generated videos, achieving a high consensus on the overall questions. More importantly, the system empowers educators to act as \textbf{"pedagogical gatekeepers,"} reclaiming agency over the AI-driven creative process. We discuss the implications of our work for the design of future AI authoring tools, focusing on the balance between productivity and agency, the necessity for adaptive content scaling to match learner demographics, and the transition from "one-shot" generation to transparent, curriculum-aligned co-creation.
Ultimately, this work seeks to foster trust and professional acceptance of generative AI in education by providing a cognitively-aware partner that ensures generated content is not only visually plausible but instructionally sound.

\section{Method}
\label{sec:method} 
\subsection{Prototype Design}

PedaCo-Gen is a human-AI collaborative authoring system designed to bridge the gap between generative AI capabilities and pedagogical efficacy.
By integrating Mayer’s Cognitive Theory of Multimedia Learning~(CTML)~\cite{mayer2009} directly into the production pipeline, the system empowers educators to produce high-quality educational videos that are both visually engaging and cognitively optimized.
A key \inj{feature} of PedaCo-Gen is its human-in-the-loop approach, which provides multiple intervention points throughout the authoring process.
\yr{By providing multiple intervention points—ranging from initial constraint setting to selective feedback application—the system mitigates the "black-box" nature of traditional Text-to-Video models.}
This \yr{human-in-the-loop approach not only ensures content accuracy and pedagogical validity but also allows educators to act as the final "pedagogical gatekeepers," filtering potential AI hallucinations~\cite{hallucination2023survey} and tailoring the output to the specific needs of their learners.}
As illustrated in Figure~\ref{fig:system}, the system architecture is structured around a \textbf{three-phase iterative workflow}.
\inj{For implementation details, please refer to Appendix~\ref{sec:appendix-implementation}}.

In the \textbf{Setup Phase}, users input learning content (Fig.~\ref{fig:system}-a) and set pedagogical constraints (Fig.~\ref{fig:system}-b) that the AI should follow when generating scripts.
While CTML constraints are pre-configured as defaults, users can add their own custom requirements.
Clicking the \texttt{Generate Script} button produces an initial draft of a pedagogy-based educational video script generated by the LLM.
The \textbf{Refinement Phase} constitutes the core of the human-AI partnership.
Users get a CTML-based review instructions, and also can ask for additional reviews tailored to their context (Fig.~\ref{fig:system}-c). 
When clicking the \texttt{Request Review} button, users receive feedback % identifying strengths and areas for improvement, 
along with a proposed revised script (Fig.~\ref{fig:system}-d).
Users can either accept all feedback by clicking the \texttt{Apply Feedback} button, or directly edit the script to selectively incorporate specific suggestions.
In the final \textbf{Output Phase}, users can view the finalized video script with scene-by-scene visual descriptions and narrations (Fig.~\ref{fig:system}-e).
Users click the \texttt{Create Video} button to generate the video and preview the result (Fig.~\ref{fig:system}-f).
A workflow progress bar (Fig.~\ref{fig:system}-g) provides visual navigation across all stages.

\subsection{Study Design}

\subsubsection{Participants}
\inj{Twenty-three education professionals participated (See Appendix~\ref{sec:appendix-user-detail} for demographics).}

\subsubsection{Tasks}
Participants performed tasks involving the generation and refinement of video scripts, followed by the evaluation of the resulting videos across three educational topics.
The topics were composed of explanation types requiring different cognitive processing: galaxy collision (causal), chaos theory (abstract concept), and DNS operation principles (procedural).
The experiment employed a within-subject design where each participant compared two conditions:
(1) Baseline: Video generation from LLM~\cite{team2023gemini} generated scripts, refined by educators with CTML guidelines provided as reference, and (2) PedaCo-Gen: Video Generation with scripts refined through our human-AI collaborative workflow.
Each video was approximately one minute long.

\subsubsection{Procedure}
This IRB-approved study was conducted via a self-paced, structured, and web-based questionnaire (approx. 30 min).
The procedure consisted of three phases:
a pre-study (5 min) for informed consent, demographics, and CTML orientation; a main experiment (20 min) involving the generation and interactive refinement of scripts, as well as the evaluation of the videos across three topics and two conditions (13 rating items per video); and a post-study (5 min) focused on system usability and open-ended feedback. Participants received a \$7 gift card upon completion.

\subsubsection{Measures}
We collected quantitative and qualitative data through (1) \textbf{CTML Evaluation:} 13 items (12 CTML principles + 1 overall validity) measured on a 5-point Likert scale and (2) \textbf{System Evaluation:} 5 items (Production Efficiency, Guide Validity, Intention to Apply in Practice, Overall Satisfaction, and Intention Reflection) and two open-ended questions for qualitative feedback. 
See Appendix~\ref{sec:appendix-measures} for details.

\subsubsection{Analysis}
For quantitative analysis, we performed \inj{Wilcoxon signed-rank tests to compare CTML scores between conditions, and Mann-Whitney U tests for between-group comparisons (e.g. gender, career stage, AI usage experience) ($\alpha = .05$)}
For qualitative data analysis, thematic analysis of open-ended responses was conducted following a six-phase framework: familiarization, coding, theme identification, review, definition, and reporting~\cite{Thematic2023}.
An inductive coding approach generated 25 initial codes.
To enhance coding consistency, the research team reached consensus through iterative discussions on discrepancies, from which major themes were derived.

\section{Findings}
Our evaluation reveals that PedaCo-Gen significantly enhances both production efficiency and pedagogical quality compared to baseline.
Educators highly valued the system’s iterative workflow for reducing trial-and-error ($M=4.26$) and the AI-driven pedagogical guidance ($M=4.04$).
Quantitative assessments showed significant improvements across all 12 CTML principles ($p < .05$), with these effects remaining consistent across gender and career stages ($p > .05$).

\subsection{User Perceptions of System Satisfaction and Efficiency}

Educators showed overall positive perception of the system's practical utility (Table~\ref{tab:system-eval}), as reflected by a high \textbf{Overall Satisfaction} score ($M = 3.78$, $SD = 1.02$).
Notably, \textbf{Production Efficiency} ($M = 4.26$, $SD = 0.90$) emerged as the most highly rated attribute.
Despite the inherent temporal cost of introducing human-in-the-loop stages, participants reported a significant reduction in the perceived effort required compared to conventional authoring methods.
This suggests that for users, efficiency is not merely a function of absolute working time but is driven by the reduction of iterative trial-and-error and increased confidence in the final output.
The \textbf{Guide Validity} ($M = 4.04, SD = 0.62$) further reinforced this sense of utility.
The remarkably low standard deviation across this metric indicates a robust consensus among participants that the CTML-based review process provides actionable and pedagogically sound feedback.
By positioning the AI as a principled "reviewer" rather than a black-box generator, PedaCo-Gen provided a structured scaffold that participants found highly relevant to their professional standards.
\yr{However, the evaluation also revealed areas for further refinement, particularly regarding \textbf{Intention to Apply in Practice} ($M = 3.91, SD = 1.32$) and \textbf{Intent Reflection} ($M = 3.78, SD = 0.83$).
While there was a general willingness to use AI-generated videos as supplementary classroom materials, the high variance in application intent highlights the influence of specific educational contexts.
For instance, in early childhood education where "hands-on, experiential learning" is the primary modality, the applicability of purely video-based content may be more constrained (\textbf{P14}).
Additionally, while the iterative revision process facilitated alignment between the AI output and the educator’s vision, the relatively lower scores in intent reflection suggest that translating abstract instructional goals into generative prompts remains a complex interaction challenge.}

\begin{table}[t]
\centering
\caption{System Usability Evaluation Results. Values represent $M$ ($SD$) on a 5-point Likert scale ($N$=23).}
\small
\setlength{\tabcolsep}{4pt}
\begin{tabular}{lccccc}
\toprule
& \makecell{Production \\ Efficiency}
& \makecell{Guide \\ Validity}
& \makecell{Intention \\ to Apply}
& \makecell{Overall \\ Satisfaction}
& \makecell{Intent \\ Reflection} \\
\midrule
$M$ ($SD$) & 4.26 (0.90) & 4.04 (0.62) & 3.91 (1.32) & 3.78 (1.02) & 3.78 (0.83) \\
\bottomrule
\end{tabular}
\label{tab:system-eval}
\end{table}

\subsection{Pedagogical Effectiveness: Video Quality Assessment Based on CTML Principles}
The pedagogical efficacy of PedaCo-Gen was evaluated through a comparative analysis of Mayer’s 12 CTML principles~\cite{mayer2009}. 
Quantitatively, the system demonstrated significant improvements across all tested educational topics—causal, abstract, and sequential—with all gains reaching statistical significance ($p < .05$). 
For a detailed breakdown of performance by topic, please refer to Appendix~\ref{sec:appendix-ctml-results}.

Across all principles, the average score increased from $3.07$ ($SD = 0.81$) to $3.86$ ($SD = 0.66$), representing a mean improvement of $+0.79$ points ($p < .01$).
The most significant gains were observed in the \textbf{Pre-training Principle} ($+0.86$) and the \textbf{Coherence Principle} ($+0.84$).
The high improvement in the \textbf{Pre-training Principle} (+0.86) indicates that the AI reviewer guided the introduction of key concepts and terms at the beginning of videos, strengthening designs that help learners form background knowledge. 
The improvement in the \textbf{Coherence Principle} (+0.84) suggests that decorative elements unrelated to learning were removed, increasing learning focus.

On the other hand, while improvements were observed across all principles, some items showed relatively lower gains.
The lower improvement in the \textbf{Spatial Contiguity Principle} (+0.53) and the \textbf{Signaling Principle} (+0.51) is related to the limitations of current Text-to-Video (T2V) models.
T2V models still tend to generate typos when rendering text, resulting in constraints on accurate placement and emphasis of visual elements and text.
These results suggest that while PedaCo-Gen's pedagogical review function is effective, improving the text rendering accuracy of the T2V model itself is necessary for ultimate quality improvement.

\section{Discussion}

\subsection{AI as a Metacognitive Scaffold: Navigating the Agency-Productivity Trade-off}
PedaCo-Gen functions as an interactive metacognitive scaffold, prompting educators to use CTML guidelines as a heuristic checklist rather than passively accepting AI outputs.
This "productive friction" is reflected in the high validity scores (4.0/5.0) and qualitative feedback.
For instance, while \textbf{P23} found \textit{"inducing consistent results [...] was quite challenging,"} the participant noted that such iterative refinement is essential for a \textit{"robust and effective learning tool".}
Furthermore, the human-in-the-loop requirement reinforced the educator’s role as a \textbf{"pedagogical gatekeeper."} 
As \textbf{P05} emphasized, the necessity for \textit{"separate functions where teachers can additionally review, edit, and modify"} highlights that the system does not replace human expertise but augments it, ensuring that the final artifact remains pedagogically sound through deliberate human oversight.

\subsection{Contextual Granularity and Curricular Alignment}
A salient theme was the necessity for \textbf{adaptive instructional tailoring} based on the learner's developmental stage.
Participants (\textbf{P01, 05, 12, 14, 20}) highlighted the critical need for modulating content difficulty based on the target audience.
\textbf{P12} pointed out that \textit{"video difficulty should vary according to the level of the target learner,"} highlighting the critical need for modulating content difficulty.
Similarly, \textbf{P01, 12, 19} noted the effectiveness for different grades, agreeing that while the visual depth seems suitable for older students, younger learners require a higher degree of \textbf{ease of understanding} through \textit{"simplified language and frequent real-world examples"}~(\textbf{P19}). 

These insights suggest that the \textbf{Personalization Principle} must extend beyond conversational tone to encompass \textbf{dynamic content scaling} and \textbf{curricular alignment}.
Future iterations should allow educators to define target learner demographics  to automatically modulate vocabulary, conceptual depth, and visual complexity.
Furthermore, as \textbf{P12} suggested, \textit{"It would be ideal to train the AI system on the official national curriculum"}, aligning the system with \textbf{formal curricular standards} would ensure that AI-generated scripts move beyond generic explanations to become context-aware, reliable instructional materials.

\subsection{Transparency and the Explainability Frontier}
While PedaCo-Gen aimed to demystify the generation process by introducing an \textbf{Intermediate Representation (IR)} phase—allowing educators to review and edit scripts before video synthesis—our findings reveal that a sense of opacity persisted for several participants.
\textbf{P12} voiced concerns over the provenance of the data, saying that \textit{"While the quality of the video is good, I have no way of knowing if the prompts are generated based on actual textbook examples or valid curriculum content."}
This highlights that for educators, transparency is not just about the process, but about the reliability of the data driving the generation.

Furthermore, the stochastic nature of generative models created a sense of "unpredictability," which participants perceived as a lack of transparency in control.
\textbf{P23} observed: \textit{"I found it quite challenging to induce consistent results even when using the script guide."}
This difficulty in achieving consistent alignment between user intent and AI output suggests that the internal transformations within the AI remain opaque, even when pedagogical scaffolds are provided. 

Also, other participant suggested that clarity in the generation process could enhance utility (\textit{"If I could understand the generation process in more detail, I would be able to utilize it more effectively across various domains."} (\yr{\textbf{P17}}).)
In an educational context, trust is predicated on the "why"—why a particular scene was generated or how it connects to formal educational standards.
Integrating \textbf{Explainable AI (XAI)~\cite{xai1, xai2}} techniques could bridge this gap.
Future work must move beyond showing just \textit{what} the AI produced to explaining \textit{how} it aligns with pedagogical constraints.
By explicitly visualizing the pedagogical intent (e.g., "This background was simplified to minimize extraneous cognitive load per the Coherence Principle"), the system could foster a more transparent and credible partnership with the educator.

\section{Limitations and Future Work}
Technical limitations also persist, particularly regarding \textbf{audio synthesis quality}. 
Participants noted that \textit{"some videos had slightly awkward audio"} (\textbf{P01}) and \textit{"the voice is too unnatural"} (\textbf{P20}).
Future work will focus on integrating high-fidelity, emotionally expressive Text-to-Speech (TTS) models.
As this study focused on the system workflow, measuring cognitive load and learning outcomes with actual learners remains for future research.\
Future studies could employ independent raters to mitigate potential self-evaluation bias.
Additionally, generalizability is limited due to the focus on Korean educators and science topics; we plan to extend validation across diverse domains and cultural contexts.
Finally, transitioning from experimental use to classroom implementation will require \textbf{comprehensive training programs} that focus on "AI literacy" for educators, moving from simple prompting to principled co-creation. 
As \textbf{P22} suggested: \textit{"I would like to have training workshops focused on practical classroom applications."}
\section{Conclusion}

This research demonstrates the efficacy of PedaCo-Gen in addressing the pedagogical shortcomings of unconstrained Text-to-Video (T2V) models. Our evaluation with 23 education experts confirms that the system achieves statistically significant quality improvements across all educational topics and CTML principles compared to baseline models. More importantly, PedaCo-Gen functions not merely as an automation tool but as an interactive metacognitive scaffold that empowers educators to critically reflect on their instructional designs while mitigating the technical burdens of video production. By facilitating an iterative review of intermediate representations, the system enables educators to reclaim their role as \textbf{"pedagogical gatekeepers,"} ensuring that the final output remains instructionally sound through deliberate human oversight.

Ultimately, PedaCo-Gen signals a paradigm shift from "one-shot" automated generation toward a model of \textbf{principled human-AI co-creation}. This work offers a roadmap for the design of future educational authoring tools that prioritize human agency and domain expertise. Future research will focus on enhancing context-aware adaptation—such as dynamic content scaling based on learner demographics—and integrating Explainable AI (XAI) to further demystify the generative process. By fostering a transparent and collaborative partnership between humans and AI, we aim to cultivate a more reliable and pedagogically grounded ecosystem for digital instructional content creation.

\begin{acks}
This work was funded by the Korean Government through the grants from IITP (RS-2021-II211343, RS-2025-25442338) and KOCCA (RS-2024-00398320).
The web-based prototype presented in this study was developed with the assistance of a Large Language Model (LLM).
\end{acks}

\bibliographystyle{ACM-Reference-Format}
\bibliography{main}

\appendix
\newpage
\section*{Appendix}

\section{Cognitive theory of multimedia learning}
\label{sec:appendix-ctml}

\begin{table*}[t]
\centering
\caption{Mayer's 12 Principles of Multimedia Learning, categorized by their primary cognitive function. These principles provide a foundational framework for designing instruction that is congruent with human cognitive architecture.}
\label{tab:mayer_principles}
\begin{tabularx}{\textwidth}{lXX}
\toprule
\textbf{Principle} & \textbf{Guideline (What to do)} & \textbf{Cognitive Rationale (Why it works)} \\
\midrule
\multicolumn{3}{l}{\textit{\textbf{I. Principles for Reducing Extraneous Processing}}} \\
\midrule
Coherence & Exclude extraneous, irrelevant material. & Reduce cognitive load by preventing distraction from non-essential information. \\
Signaling & Highlight essential information. & Direct the learner's limited attention to critical elements. \\
Redundancy & Avoid presenting identical information simultaneously in text and narration. & Prevent overload from processing redundant verbal information in two channels. \\
Spatial Contiguity & Place corresponding words and pictures near each other on the screen. & Reduce the cognitive effort needed to mentally integrate related information. \\
Temporal Contiguity & Present corresponding words and pictures at the same time. & Reduce the cognitive load of holding information in working memory while waiting for the other part. \\
\midrule
\multicolumn{3}{l}{\textit{\textbf{II. Principles for Managing Essential Processing}}} \\
\midrule
Segmenting & Break the lesson into smaller, learner-paced segments. & Help manage the complexity of the material by allowing learners to process one at a time. \\
Pre-training & Introduce key concepts and their names before the lesson. & Activate relevant prior knowledge and reduce the load during the main lesson. \\
Modality & Present words as narration rather than on-screen text, especially for complex visuals. & Distribute cognitive processing across both visual and auditory channels, avoiding overload in the visual channel. \\
\midrule
\multicolumn{3}{l}{\textit{\textbf{III. Principles for Fostering Generative Processing}}} \\
\midrule
Multimedia & Present information using both words and pictures rather than words alone. & Encourage learners to build connections between visual and verbal mental models. \\
Personalization & Use a conversational and informal tone. & Promote social engagement, which encourages deeper cognitive processing. \\
Voice & Use a human voice for narration rather than a machine voice. & A human voice can better convey social cues so that learners engage with the material. \\
Image & Use clear, high-quality visuals that are directly relevant to the content, and avoid extraneous or technically poor images (e.g., an unnecessary "talking head"). & Ensure the learner's cognitive resources are focused on understanding the content, rather than being diverted by processing irrelevant social cues (from an instructor's image) or deciphering technically poor visuals. \\
\bottomrule
\end{tabularx}
\end{table*}

Table.~\ref{tab:mayer_principles} shows the Mayer's 12 principles of multimedia learning from Cognitive Theory of Multimedia Learning~(CTML)~\cite{mayer2009}.
\section{Measures}
\label{sec:appendix-measures}

This appendix presents the questionnaire items used in the user study. Pre-survey questions (Table~\ref{tab:pre-survey}) collected participants' demographic information and prior AI experience. As described in Section~\ref{sec:method}, CTML-based video quality evaluation items (Table~\ref{tab:ctml-items}) assessed the generated videos under Baseline and PedaCo-Gen conditions for each of three topics, according to multimedia learning principles. System usability evaluation items (Table~\ref{tab:system-survey}) measured participants' perceptions of the PedaCo-Gen system. All questionnaire items were presented in the participants' native language (Korean), and the English translations provided here were generated using the DeepL translation service.

\begin{table*}[t]
\centering
\caption{Pre-survey Questions}
\label{tab:pre-survey}
\begin{tabular}{c|p{0.60\textwidth}}
\hline
\textbf{Question ID} & \textbf{Questions} \\
\hline
Q1 & What is your current occupation (or major)? \\ \hline
Q2 & How much education-related experience do you have (total years of work and study combined)? \\ \hline
Q3 & How often do you use generative AI (ChatGPT, Claude, Sora, etc.)? \\ \hline
Q4 & Have you ever used AI-generated materials as supplementary resources in actual educational settings (classroom, lectures, etc.)? \\ \hline
Q5 & Please describe specifically how you created those supplementary materials. \\ \hline
Q6 & What is the reason for not creating/using supplementary materials with AI? \\
\hline
\end{tabular}
\end{table*}
 
\begin{table*}[h]
\centering
\caption{CTML-based Video Quality Evaluation Items: 12 CTML principles + 1 Overall Validity (13 items × 2 conditions × 3 topics = 78 items per participant)}
\label{tab:ctml-items}
\begin{tabular}{c|p{0.60\textwidth}|l}
\hline
\textbf{Question ID} & \textbf{Questions} & \textbf{Principles} \\
\hline
Q1 & Are images/videos appropriately combined to aid understanding of the learning content, rather than providing text (language) alone? & Multimedia \\ \hline
Q2 & Are distracting backgrounds, unnecessary sound effects, and decorative elements unrelated to the learning content removed to maintain high learning concentration? & Coherence \\ \hline
Q3 & Are key words or important visual elements clearly guided through subtitles, arrows, highlights, etc.? & Signaling \\ \hline
Q4 & When complex graphics and text appear on screen simultaneously, does the narration focus on explaining the graphics rather than simply reading the on-screen text? & Redundancy \\ \hline
Q5 & Are explanatory text and related images placed close together to prevent visual distraction? & Spatial Contiguity \\ \hline
Q6 & Are narration (explanations) and corresponding visual materials presented simultaneously without time gaps? & Temporal Contiguity \\ \hline
Q7 & Is the content divided into appropriate units (steps) that learners can digest, rather than being presented all at once? & Segmenting \\ \hline
Q8 & Before the main explanation, are key terms and concept characteristics defined or introduced in advance to build background knowledge? & Pre-training \\ \hline
Q9 & Is information conveyed through harmonious use of animation and narration, rather than simply listing text on screen? & Modality \\ \hline
Q10 & Does the writing style use friendly, conversational expressions that speak to learners, rather than overly formal language? & Personalization \\ \hline
Q11 & Does the narration voice convey natural intonation and emotion like a human voice, rather than sounding mechanical? & Voice \\ \hline
Q12 & Is the screen quality high enough to allow learners to focus on the learning content (graphics) itself? & Image \\ \hline
Q13 & Was the video content generated by the system overall appropriate and valid from the perspective of learning content? & Overall Validity \\
\hline
\end{tabular}
\end{table*}

\begin{table*}[h]
\centering
\caption{System Usability Evaluation Items}
\label{tab:system-survey}
\begin{tabular}{c|p{0.60\textwidth}|l}
\hline
\textbf{Question ID} & \textbf{Questions} & \textbf{Type} \\
\hline
Q1 & Were you satisfied with your overall experience using the PedaCo-Gen system? & Overall Satisfaction \\ \hline
Q2 & Was the pedagogical review (CTML guide questions 1-12) proposed by the system appropriate and valid from an education expert's perspective? & Guide Validity \\ \hline
Q3 & Was your educational intent sufficiently reflected in the final video through the iterative revision (Review \& Regenerate) process? & Intent Reflection \\ \hline
Q4 & Would using this system significantly reduce the time and effort required to produce high-quality educational videos compared to conventional methods? & Production Efficiency \\ \hline
Q5 & Would you be willing to use videos generated by this AI system as supplementary materials in actual educational settings (classroom, lectures, etc.)? & Intention to Apply \\ \hline
Q6 & If you are not willing to use the AI generation system as supplementary material, what improvements would make you willing to use it? & Open-ended \\ \hline
Q7 & Please freely describe your opinions about the AI generation system used in this survey. & Open-ended \\
\hline
\end{tabular}
\end{table*}

\section{Participant Demographics}
\label{sec:appendix-user-detail}

Initially, a total of 24 participants were recruited for the study.
However, one participant was excluded due to an insincere response pattern (straight-lining with repeated identical responses).
Consequently, data from 23 participants were used in the final analysis (14 female, 9 male).
The mean age of all participants was 31.3 years ($SD = 8.8$), with male participants averaging 34.6 years ($SD = 8.0$) and female participants averaging 29.1 years ($SD = 8.7$).
Detailed demographic information is presented in Table~\ref{tab:demographics}.

\begin{table*}[t]
    \centering
    \caption{Demographic Information of Study Participants.}
    \label{tab:demographics}

    \begin{tabular}{ccccccc}
        \toprule
        \textit{ID} & \textit{Gender} & \textit{Age} & \textit{Occupation} & \textit{Experience} & \textit{AI Usage Freq.} & \textit{Prior Classroom AI Use} \\
        \midrule
        P01 & F & 27 & Elementary Teacher & 3+ years & 1-2/week & No \\
        P02 & F & 27 & Middle School Teacher & 3+ years & $\geq$3/week & Yes \\
        P03 & F & 27 & High School Teacher & 3+ years & $\geq$3/week & Yes \\
        P04 & F & 23 & Education Major & 3+ years & 1-2/week & Yes \\
        P05 & F & 22 & Early Childhood Teacher & 3+ years & 1-2/week & Yes \\
        P06 & M & 27 & Elementary Teacher & 3+ years & $\geq$3/week & Yes \\
        P07 & M & 30 & High School Teacher & 5+ years & 1-2/week & Yes \\
        P08 & F & 29 & Middle School Teacher & 5+ years & $\geq$3/week & Yes \\
        P09 & F & 25 & Elementary Teacher & 1+ years & $\geq$3/week & Yes \\
        P10 & M & 27 & Education Major & 5+ years & 1-2/week & Yes \\
        P11 & F & 26 & High School Teacher & 3+ years & 1-2/month & Yes \\
        P12 & F & 26 & Middle School Teacher & 10+ years & $\geq$3/week & Yes \\
        P13 & F & 27 & High School Teacher & 1+ years & $\geq$3/week & Yes \\
        P14 & F & 57 & Early Childhood Teacher & 10+ years & Rarely & No \\
        P15 & F & 24 & Education Major & 5+ years & $\geq$3/week & Yes \\
        % PXX & F & 38 & Early Childhood Teacher & 5+ years & $\geq$3/week & No \\
        P16 & F & 40 & Early Childhood Teacher & 10+ years & 1-2/week & No \\
        P17 & M & 55 & Education Major & 10+ years & 1-2/week & Yes \\
        P18 & M & 37 & Middle School Teacher & 10+ years & 1-2/month & Yes \\
        P19 & F & 28 & High School Teacher & 3+ years & $\geq$3/week & Yes \\
        P20 & M & 33 & High School Teacher & 5+ years & 1-2/week & Yes \\
        P21 & M & 35 & Middle School Teacher & 5+ years & 1-2/week & Yes \\
        P22 & M & 32 & High School Teacher & 5+ years & 1-2/week & No \\
        P23 & M & 35 & Middle School Teacher & 5+ years & $\geq$3/week & Yes \\
        \bottomrule
    \end{tabular}
\end{table*}

\section{Results of CTML-based Video Assessment}
\label{sec:appendix-ctml-results}

Table~\ref{tab:ctml_improvement} presents the improvement in video quality ratings for each CTML principle when comparing the Baseline and PedaCo-Gen conditions. A Wilcoxon signed-rank test was used for statistical analysis ($\alpha = .05$). All 13 evaluation items showed statistically significant improvements, with the largest gains observed in Overall Validity (+0.96), Pre-training (+0.86), and Coherence (+0.84). Effect sizes were measured using rank-biserial correlation $r$ (small: $r \geq .1$, medium: $r \geq .3$, large: $r \geq .5$).

\begin{table*}[h]
\centering
\caption{Improvement by CTML Principle (Averaged Across All Topics)}
\label{tab:ctml_improvement}
\begin{tabular}{clccccccc}
\toprule
Rank & CTML Principle & Baseline & PedaCo-Gen & Improvement & $p$-value & Sig. & $r$ & Effect Size \\
\midrule
1 & Overall Validity & 3.07 & 4.03 & +0.96 & $< .01$ & * & .88 & large \\
2 & Pre-training & 2.84 & 3.70 & +0.86 & $< .01$ & * & .78 & large \\
3 & Coherence & 3.03 & 3.87 & +0.84 & $< .01$ & * & .96 & large \\
4 & Personalization & 3.22 & 3.97 & +0.75 & $< .01$ & * & .82 & large \\
5 & Multimedia & 3.33 & 3.99 & +0.66 & $< .01$ & * & .79 & large \\
6 & Segmenting & 3.51 & 4.17 & +0.66 & $< .01$ & * & .80 & large \\
7 & Voice & 3.07 & 3.70 & +0.63 & $< .01$ & * & .70 & large \\
8 & Modality & 3.41 & 4.00 & +0.59 & $< .01$ & * & .71 & large \\
9 & Image & 3.45 & 4.04 & +0.59 & $< .01$ & * & .83 & large \\
10 & Temporal Contiguity & 3.62 & 4.20 & +0.58 & $< .01$ & * & .61 & large \\
11 & Spatial Contiguity & 3.14 & 3.67 & +0.53 & $< .01$ & * & .69 & large \\
12 & Signaling & 3.00 & 3.51 & +0.51 & $< .01$ & * & .64 & large \\
13 & Redundancy & 3.33 & 3.74 & +0.41 & $.045$ & * & .42 & medium \\
\bottomrule
\end{tabular}
\end{table*}

Table~\ref{tab:topic_comparison} presents the comparison of CTML Overall Validity scores by topic. A Wilcoxon signed-rank tests were used for statistical analysis ($\alpha = .05$). Statistically significant improvements were observed across all topics.

\begin{table*}[h]
\centering
\caption{Comparison of CTML Overall Validity (Q13) by Topic. Topic 1: causal explanation (galaxy collision), Topic 2: abstract concept explanation (chaos theory), Topic 3: sequential explanation (DNS operation).}
\label{tab:topic_comparison}
\begin{tabular}{llccclc}
\toprule
Topic & Explanation Type & Baseline & PedaCo-Gen & Improvement & $p$-value & Sig. \\
\midrule
Topic 1 & Causal & 3.26 & 4.13 & +0.87 & $< .01$ & * \\
Topic 2 & Abstract concept & 2.74 & 3.91 & +1.17 & $< .01$ & * \\
Topic 3 & Sequential & 3.22 & 4.04 & +0.82 & $< .01$ & * \\
\bottomrule
\end{tabular}
\end{table*}

The largest improvement (+1.17) in Topic 2 (abstract concept explanation) suggests that the AI reviewer's structured feedback was particularly effective for content dealing with abstract concepts. This can be interpreted as the importance of visualization and explanation structure being more pronounced for abstract concepts, where learners have difficulty forming concrete mental images.
\section{Subgroup Analysis Results}
\label{sec:appendix-subgroup}

\subsection{Differences by AI Tool Usage Experience}

No statistically significant differences were found in system evaluation items based on AI tool usage experience ($p > .05$).

However, qualitative responses revealed perceptual differences according to AI tool usage experience. One participant noted that \textit{``for those who have used AI tools like ChatGPT, Gemini, Gamma, or Vrew... it seems a bit frustrating or rather cumbersome,''} suggesting that users familiar with AI tools may perceive PedaCo-Gen's structured workflow as restrictive. While general-purpose AI tools allow free-form prompting, PedaCo-Gen requires a step-by-step review process based on pedagogical principles. Conversely, for users with less AI tool experience, this guided structure may lower the barrier to entry.

As \textbf{P14} noted, this relates to the perceived complexity of the tool: \textit{``Both the AI and the methods for using it are far too difficult; I even find general computer usage challenging.''}

\subsection{Differences by Gender and Career}

The participant composition was 9 males (39.1\%) and 14 females (60.9\%). No statistically significant differences were found between genders in system evaluation items ($p > .05$). Females ($M = 4.00$) showed a higher tendency than males ($M = 3.14$) in intent reflection, but this did not reach statistical significance ($p = .053$).

No statistically significant differences were found between career groups in CTML evaluation scores or system evaluation items ($p > .05$). These results suggest that the PedaCo-Gen system provides consistent effects across education professionals regardless of gender and career experience.

\section{Implementation Details of PedaCo-Gen}
\label{sec:appendix-implementation}

This appendix describes the system implementation and prompt design of PedaCo-Gen. We first present an overview of the platform's core functions and user workflow, followed by the prompts used to guide the large language model in generating and reviewing educational video scripts. These prompts are provided as default settings, and educators can customize them by adding their own principles, constraints, or output formats to better align with their specific pedagogical goals.

\subsection{System Implementation}
Our platform utilizes Gemini~\cite{team2023gemini} 2.5 Flash for script generation and CTML-based review, while Veo~\cite{google2024veo} 3.1 handles video synthesis, performing three core functions.
First, the script generation function creates educational video scripts by applying CTML principles and constraints to the learning content input by users. Scripts are structured by scene, with each scene containing visual descriptions and audio narration.
The number of scenes is automatically determined by the LLM based on content, though users can also configure this manually.
Second, the CTML-based review function assigns the LLM the role of `an expert reviewer who meticulously examines and provides feedback on educational video scripts,' comprehensively evaluating the 12 CTML principles. Feedback is output as improvement suggestions along with a proposed revised script.
Third, the video synthesis function passes the final script to Veo 3.1 scene by scene, where videos are generated individually and then combined.

To begin, users input their learning content by copying and pasting well-structured text knowledge into the platform.
During the script editing process, users can either directly apply the LLM's suggestions or manually edit the text input field.
Since the review function outputs structured feedback, users can apply the suggested revised script entirely with a single ``Apply Review'' button, or selectively incorporate changes by manually editing the text field.
The review process can be repeated until the user is satisfied with the script quality.
For video synthesis, the default duration per scene is configured to 8 seconds to match Veo 3.1's maximum output length; users can adjust this setting when using alternative video generation models.

\subsection{Script Generation Stage}
\label{sec:generation-prompt}

Table~\ref{tab:generation_prompt} presents the full prompt provided to the LLM for educational video script generation.

\begin{table*}[h]
\centering
\caption{Full LLM Prompt for Script Generation.}
\label{tab:generation_prompt}
\begin{tabularx}{\textwidth}{X}
\toprule
\textbf{Full LLM Prompt for Script Generation} \\
\midrule
\textit{\textbf{[Principles and Constraints for Educational Video Production]}} \\
\midrule
\textbf{<Principles>}
\begin{enumerate}[label=\arabic*., wide, labelindent=0pt]
    \item \textbf{Coherence}
        \begin{itemize}[label=–, leftmargin=*]
            \item Background music should not be used.
            \item The learning content must be preserved in both the narration and the video.
            \item The content of the video and the learning content must be directly related.
            \item The content of the narration and the learning content must be directly related.
        \end{itemize}
    \item \textbf{Modality \& Redundancy}
        \begin{itemize}[label=–, leftmargin=*]
            \item Use images or voice-over narration instead of on-screen text.
            \item Educational content included in the narration should have minimal corresponding text displayed on the screen.
        \end{itemize}
    \item \textbf{Learner-Friendly}
        \begin{itemize}[label=–, leftmargin=*]
            \item The narration script should be written in a friendly and gentle conversational style.
            \item Use a first-person, informal, and conversational tone.
            \item Use a standard human voice rather than a machine voice.
        \end{itemize}
    \item \textbf{Contiguity}
        \begin{itemize}[label=–, leftmargin=*]
            \item Write the script so that narration and visuals are synchronized in time and aligned in meaning.
            \item Place related text and graphics close to each other on the screen.
        \end{itemize}
    \item \textbf{Visuals}
        \begin{itemize}[label=–, leftmargin=*]
            \item Descriptions of video scenes should be clear.
            \item Only describe scenes that directly aid in understanding the learning content; exclude decorative or irrelevant visuals.
            \item Use signaling cues (arrows, highlight colors, bold text, etc.) to direct attention to important information.
            \item Avoid displaying the speaker's face continuously; prioritize visuals that explain the content.
        \end{itemize}
    \item \textbf{Learning Flow}
        \begin{itemize}[label=–, leftmargin=*]
            \item Avoid presenting too much information in a single scene; spread it out over multiple scenes.
            \item Introduce key terms and concepts early (e.g., in Scene 1-2) before presenting complex content.
        \end{itemize}
\end{enumerate} \\
\midrule
\textbf{<Constraints>}
\begin{enumerate}[label=\arabic*., wide, labelindent=0pt]
    \item Assign a suitable length of narration to a scene.
    \item Maximum scene count: Make your own judgment.
\end{enumerate} \\
\midrule \midrule
\textit{\textbf{[Output Format]}} \\
\midrule
<Scene 1> \newline
Visual Description: ... \newline
Clear Narration: ... \newline
\newline
<Scene N> \newline
Visual Description: ... \newline
Clear Narration: ... \\
\midrule \midrule
\textit{\textbf{[Learning Content]}} \\
\midrule
<Insert the learning content here.> \\
\bottomrule
\end{tabularx}
\end{table*}

\subsection{Script Review Stage}
\label{sec:review-prompt}

Table~\ref{tab:review_prompt} presents the full prompt provided to the LLM for script review.

\begin{table*}[h]
\centering
\caption{Full LLM Prompt for Script Review.}
\label{tab:review_prompt}
\begin{tabularx}{\textwidth}{X}
\toprule
\textbf{Full LLM Prompt for Script Review} \\
\midrule
You are an expert reviewer who meticulously examines and provides feedback on educational video scripts. Referring to all the instructions below, review the provided [Video Generation Script] to ensure it accurately reflects the [Learning Content] and complies with all [Constraints and Principles]. After a detailed review, write a revised script. \\
\midrule \midrule
\textit{\textbf{[Review Criteria]}} \\
\midrule
\textbf{<Principles>}
\begin{enumerate}[label=\arabic*., wide, labelindent=0pt]
    \item \textbf{Coherence}
        \begin{itemize}[label=–, leftmargin=*]
            \item Background music should not be used.
            \item The learning content must be preserved in both the narration and the video.
            \item The content of the video and the learning content must be directly related.
            \item The content of the narration and the learning content must be directly related.
        \end{itemize}
    \item \textbf{Modality \& Redundancy}
        \begin{itemize}[label=–, leftmargin=*]
            \item Use images or voice-over narration instead of on-screen text.
            \item Educational content included in the narration should have minimal corresponding text displayed on the screen.
        \end{itemize}
    \item \textbf{Learner-Friendly}
        \begin{itemize}[label=–, leftmargin=*]
            \item The narration script should be written in a friendly and gentle conversational style.
            \item Use a first-person, informal, and conversational tone.
            \item Use a standard human voice rather than a machine voice.
        \end{itemize}
    \item \textbf{Contiguity}
        \begin{itemize}[label=–, leftmargin=*]
            \item Write the script so that narration and visuals are synchronized in time and aligned in meaning.
            \item Place related text and graphics close to each other on the screen.
        \end{itemize}
    \item \textbf{Visuals}
        \begin{itemize}[label=–, leftmargin=*]
            \item Descriptions of video scenes should be clear, specific, and of professional quality.
            \item Only describe scenes that directly aid in understanding the learning content; exclude decorative or irrelevant visuals.
            \item Use signaling cues (arrows, highlight colors, bold text, etc.) to direct attention to important information.
            \item Avoid displaying the speaker's face continuously; prioritize visuals that explain the content.
        \end{itemize}
    \item \textbf{Learning Flow}
        \begin{itemize}[label=–, leftmargin=*]
            \item Avoid presenting too much information in a single scene; spread it out over multiple scenes.
            \item Introduce key terms and concepts early (e.g., in Scene 1-2) before presenting complex content.
        \end{itemize}
\end{enumerate} \\
\midrule
\textbf{<Constraints>}
\begin{enumerate}[label=\arabic*., wide, labelindent=0pt]
    \item Assign only one narration sentence to a single scene.
    \item Maximum scene count: Make your own judgment.
\end{enumerate} \\
\midrule \midrule
\textit{\textbf{[Output Format]}} \\
\midrule
\textbf{Detailed Review Results}: \newline
\textbf{Suggestions for Improvement}: (Point out specific scene numbers where the learning content is inadequately reflected or where principles are violated, and propose clear revision plans.) \newline
\textbf{Revised Script}: (Output the entire final script reflecting all the suggested improvements.) \\
\midrule \midrule
\textit{\textbf{[Learning Content]}} \\
\midrule
<Insert the learning content here.> \\
\midrule \midrule
\textit{\textbf{[Video Generation Script]}} \\
\midrule
<Insert the video generation script here.> \\
\bottomrule
\end{tabularx}
\end{table*}

\end{document}